\documentclass[11pt,a4paper]{article}

\usepackage[utf8]{inputenc}
\usepackage[T1]{fontenc}
\usepackage[margin=1in]{geometry}
\usepackage{hyperref}
\usepackage{url}
\usepackage{booktabs}
\usepackage{multirow}
\usepackage{graphicx}
\usepackage{amsmath}
\usepackage{xspace}
\usepackage{tipa}
\usepackage{natbib}
\usepackage{parskip}
\usepackage{titlesec}
\usepackage{caption}
\usepackage[table]{xcolor}
\usepackage{threeparttable}
\usepackage{setspace}

\hypersetup{
  colorlinks=true,
  linkcolor=blue!60!black,
  citecolor=blue!60!black,
  urlcolor=blue!60!black
}

\title{\textbf{From Speech to Text Corpora: Evaluating ASR-Based Data Acquisition\\
for Low-Resource Fongbe and Hausa}}

\author{
  Mahounan Pericles Adjovi$^{1}$ \quad
  Victor Olufemi$^{1}$ \quad
  Roald Eiselen$^{1,2}$ \quad
  Prasenjit Mitra$^{1}$\\[4pt]
  $^{1}$Carnegie Mellon University Africa, Kigali, Rwanda\\
  $^{2}$North-West University, South Africa\\[4pt]
  \texttt{madjovi@alumni.cmu.edu} \quad
  \texttt{Roald.Eiselen@nwu.ac.za} \quad
  \texttt{prasenjitm@andrew.cmu.edu}
}

\date{}

\begin{document}

\maketitle

\begin{abstract}
Low-resource African languages lack text corpora needed for language model training. We investigate whether ASR pipelines can extend text resources for two typologically distinct West African languages: Fongbe (tonal, diacritic-rich) and Hausa (non-tonal). We fine-tune MMS-300M on a curated 12.3-hour Fongbe dataset, achieving 9.48\% Word Error Rate (WER) on the ALFFA benchmark---a 78\% relative reduction from the prior 44.04\% baseline---while preserving tonal diacritics critical to the language. For Hausa, we apply an existing fine-tuned Whisper-Small model. We catalog 1,553 YouTube videos (236 hours) and process a subset of 424 videos (45.49 hours) selected to balance domain diversity with available computational resources, producing 6,770 transcribed segments. Human evaluation on 50 randomly sampled segments per language shows mean quality scores of 57.4/100 for Hausa and 36.5/100 for Fongbe, indicating that while Hausa transcriptions approach acceptable quality for corpus construction, Fongbe transcriptions require post-processing or improved models for production use. We release the curated dataset, fine-tuned model, transcribed corpus, and full video catalog following platform terms and ethical guidelines.
\end{abstract}

% ============================================================
\section{Introduction}
\label{sec:introduction}
% ============================================================

The development of large language models (LLMs) depends on large-scale text corpora. For most African languages, such corpora do not exist. Fongbe, spoken by approximately 2.2 million people primarily in Benin \citep{ethnologue_fongbe}, has limited publicly available transcribed speech: the ALFFA corpus \citep{alffa} contributes roughly 6 hours of read speech with Fongbe transcriptions. While the FFSTC \citep{ffstc} (31h) and FFSTC-2 \citep{ffstc2} (61h) corpora provide substantial Fongbe audio, these are speech translation datasets pairing Fongbe audio with French translations rather than Fongbe transcriptions, limiting their utility for monolingual ASR training. Hausa, with over 80 million speakers across West Africa \citep{ethnologue_hausa}, is better resourced but still far behind high-resource languages \citep{imam2025asr}. This scarcity blocks the development of language technologies for these communities.

ASR offers a path forward: spoken content in these languages is far more abundant than written text. YouTube alone hosts thousands of videos in Fongbe and Hausa spanning news, education, music, and cultural programming. If ASR models can transcribe this audio at acceptable quality, spoken media becomes a scalable text source for corpus construction.

However, ASR for low-resource African languages remains challenging. Naive application of general-purpose multilingual systems fails: Whisper lacks Fongbe support entirely, defaulting to French transcription when encountering Fongbe's tonal diacritics and 69-character alphabet \citep{whisper}. For Hausa, despite basic Whisper support, we observe frequent misclassification as Arabic on musical and religious content due to phonetic overlap and Arabic script influence in written Hausa. Fongbe's tonal system requires diacritic preservation---a feature most multilingual ASR systems handle poorly. Code-switching with French (for Fongbe) and English (for Hausa) introduces systematic transcription errors. While recent models---MMS \citep{mms}, AfriHuBERT \citep{afrihubert}---and datasets enabling model development such as FFSTC-2 \citep{ffstc2} have improved performance on curated laboratory recordings, their behavior on uncurated, in-the-wild YouTube audio with background noise, music, and domain variation remains largely uncharacterized.

We ask: \textbf{Can ASR models trained on curated speech corpora produce usable transcriptions from in-the-wild YouTube audio to expand text resources for low-resource Fongbe and Hausa?}

We answer this question through a three-stage investigation: (1)~curating a 12.3-hour Fongbe dataset and fine-tuning MMS-300M to 9.48\% WER, a 78\% relative reduction from the 44.04\% baseline; (2)~systematically cataloging 1,553 YouTube videos (236 hours) across both languages; (3)~processing a subset of 424 videos (45.49 hours) selected to balance domain coverage with computational constraints, producing 6,770 transcribed segments. Human evaluation on 50 samples per language reveals a performance gap: Hausa transcriptions achieve mean quality of 57.4/100 (approaching usable quality), while Fongbe scores 36.5/100, indicating that while 12 hours suffice to train functional ASR for curated data (9.48\% WER), additional work is needed for reliable in-the-wild transcription of tonal languages.

\smallskip
\noindent\textbf{Contributions.}
\begin{enumerate}
  \item A fine-tuned MMS-300M Fongbe ASR model achieving 9.48\% WER with diacritic preservation, the best reported result on the ALFFA benchmark under comparable evaluation conditions (Section~\ref{sec:results}).
  \item A curated 12.3-hour tone-preserved Fongbe speech dataset merging ALFFA and Zenodo sources (Section~\ref{sec:methodology}).
  \item A comprehensive YouTube video catalog of 1,553 videos (236 hours) with transcriptions for a processed subset of 424 videos yielding 6,770 segments (Section~\ref{sec:results}).
  \item Analysis demonstrating that 12 hours of curated data suffice to achieve competitive ASR performance on extremely low-resource tonal languages, with implications for corpus expansion strategies (Section~\ref{sec:discussion}).
\end{enumerate}
All resources and documentation will be released upon acceptance.\footnote{Project documentation: \url{https://asr-fongbe-hausa-16dz.vercel.app/}.}

% ============================================================
\section{Related Work}
\label{sec:related}
% ============================================================

\paragraph{ASR for African languages.}
Recent surveys \citep{imam2025asr,imam2025asr_slr} document the growing but still limited ASR landscape for African languages. Meta's MMS \citep{mms} extended ASR coverage to 1,100+ languages, reducing WER by more than half compared to prior multilingual systems, and remains one of few models with Fongbe support through its CTC-based architecture. OpenAI's Whisper \citep{whisper}, trained on 680K hours of weakly supervised speech, supports Hausa directly but lacks Fongbe coverage entirely, defaulting to French when encountering Fongbe audio. AfriHuBERT \citep{afrihubert}, an Africa-centric self-supervised model pretrained on 10K+ hours across 1,226 African languages, improved language identification by 3.6\% F1 and ASR by $-$2.1\% WER over its mHuBERT-147 baseline, demonstrating the value of region-specific pretraining. Wav2Vec2 and its cross-lingual variant XLS-R \citep{babu2021xlsr} have been widely adopted for low-resource ASR fine-tuning, with language-specific checkpoints now available for both Hausa and Fongbe. Despite this progress, all of these systems have been evaluated primarily on curated benchmarks; their performance on uncurated, in-the-wild audio remains largely uncharacterized.

\paragraph{Fongbe and Hausa resources.}
For Fongbe, the FFSTC \citep{ffstc} (31h) and FFSTC-2 \citep{ffstc2} (61h) corpora represent the most substantial public speech resources, providing Fongbe audio paired with French text translations for speech translation. FFSTC-2 demonstrated that tone-marking significantly affects both ASR and downstream translation quality. The ALFFA corpus \citep{alffa} provides approximately 6 hours of read Fongbe speech with Fongbe transcriptions; \citet{laleye2016fongbe} established a baseline WER of 44.04\% using Kaldi-based GMM/HMM systems on this dataset in 2016. For Hausa, Mozilla Common Voice \citep{commonvoice} provides 195K+ crowd-sourced samples, NaijaVoices \citep{naija_voices} contributes approximately 600 hours of Nigerian speech, and dedicated Whisper fine-tunes (e.g., NCAIR Hausa-ASR \citep{ncair_hausa}) have been developed by government institutions. These curated resources contrast sharply with the uncurated YouTube content we target in this work.

\paragraph{African language NLP.}
The broader challenge of building NLP resources for African languages has been addressed through both targeted dataset creation and model development. AfriBERTa \citep{afroberta} showed that encoder models pretrained on fewer than 1GB of African language data can compete with mBERT. AfroXLMR \citep{afroxlmr} improved downstream task performance through Africa-specific continual pretraining of XLM-R. InkubaLM \citep{inkubalm}, a compact 0.4B-parameter model trained on five African languages including Hausa, demonstrated competitive performance with larger multilingual models. On the annotation side, MasakhaNER \citep{masakhaner} and MasakhaPOS \citep{masakhapos} provide named entity and part-of-speech annotations for 20 languages including both Fongbe and Hausa. All of these efforts underscore that text data scarcity remains the primary bottleneck for African language NLP.

\paragraph{ASR-based corpus expansion.}
This approach is well established for high-resource languages, where semi-supervised and self-training pipelines routinely generate millions of pseudo-labeled utterances. However, applying this paradigm to low-resource African languages introduces unique challenges: language identification is unreliable for unsupported languages like Fongbe, code-switching with colonial languages (French, English, Arabic) introduces systematic transcription errors, and the quality gap between curated laboratory recordings and in-the-wild media content is substantially larger than in high-resource settings. Our work provides the first end-to-end evaluation of ASR-based corpus expansion for Fongbe and Hausa, spanning data curation, model fine-tuning, and large-scale YouTube transcription.

% ============================================================
\section{Methodology}
\label{sec:methodology}
% ============================================================

\subsection{Curated Fongbe Speech Dataset}

We constructed a unified Fongbe dataset by merging two open sources: (1)~the ALFFA corpus \citep{alffa} (Kaldi-formatted read speech) for train and test splits, and (2)~the Zenodo Fongbe Speech Dataset \citep{zenodo_fongbe} (crowd-sourced recordings with speaker metadata) for validation. All audio was resampled to 16\,kHz. Transcriptions preserve Fongbe-specific characters ({\textipa{\textsubarch{d}}}, {\textipa{E}}, {\textipa{O}}) and tonal diacritics via NFD/NFC normalization. We verified zero filename overlap between splits to prevent data leakage.

\begin{table}[ht]
  \centering
  \caption{Curated Fongbe speech dataset.}
  \label{tab:fon_dataset}
  \begin{tabular}{lrrr}
    \toprule
    \textbf{Split} & \textbf{Samples} & \textbf{Duration} & \textbf{Source} \\
    \midrule
    Train       & 8,234  & 5.73\,h & ALFFA \\
    Validation  & 3,179  & 5.11\,h & Zenodo \\
    Test        & 2,168  & 1.45\,h & ALFFA \\
    \midrule
    \textbf{Total} & \textbf{13,581} & \textbf{12.29\,h} & \\
    \bottomrule
  \end{tabular}
\end{table}

The test set is identical to that used in prior work \citep{laleye2016fongbe}, enabling direct comparison with the 44.04\% baseline.

\subsection{YouTube Video Dataset}

We systematically cataloged 1,553 YouTube videos through manual playlist curation and automated keyword search via the YouTube Data API v3 (423 Fongbe, 1,130 Hausa; 236 hours total across 761 channels). From this catalog, we selected 424 videos for download and transcription based on available computational resources and dataset construction priorities. Video selection emphasized domain diversity (educational, news, cultural, and music content) and prioritized channels with clearer audio quality based on manual inspection of sample segments. The remaining catalog videos are available for future transcription as computational resources allow. Table~\ref{tab:youtube_data} summarizes the processed subset.

Audio was extracted using \texttt{yt-dlp}, converted to mono 16\,kHz WAV, and segmented into 20--25\,s chunks with 1\,s overlap for context preservation.

\begin{table}[ht]
  \centering
  \caption{YouTube video dataset (transcribed subset, 424 videos).}
  \label{tab:youtube_data}
  \begin{tabular}{lrr}
    \toprule
    \textbf{Language} & \textbf{Hours} & \textbf{Domains} \\
    \midrule
    Fongbe & 24.91 & Education, Music, Culture \\
    Hausa  & 20.57 & News, Music, Culture \\
    \midrule
    \textbf{Total} & \textbf{45.49} & \\
    \bottomrule
  \end{tabular}
\end{table}

The Fongbe subset includes educational channels (\emph{Apprendre le Fongb\'{e}}, \emph{WA KPLON FONGBE}: 29 and 35 videos respectively) and cultural/music content (\emph{louange fon-goun}: 47 videos). The Hausa subset spans music (\emph{Hausa songs}: 85 videos), news (BBC Hausa, VOA Hausa: 112 videos), and entertainment content.

\subsection{ASR Models}

\paragraph{Fongbe.} We fine-tuned \texttt{facebook/mms-300m} \citep{mms} on our curated dataset using CTC loss, learning rate $1\times10^{-4}$, effective batch size 64, AdamW optimizer, mixed precision (FP16), for 30 epochs on an NVIDIA H100 GPU. Training converged after approximately 2 hours.

\paragraph{Hausa.} We used NCAIR Hausa-ASR \citep{ncair_hausa}, a Whisper-Small \citep{whisper} model fine-tuned for Hausa by Nigeria's National Centre for Artificial Intelligence and Robotics, applied without additional fine-tuning. We selected this model based on its reported performance on Common Voice Hausa and institutional backing.

\subsection{Large-Scale Inference}

The 424 selected videos were processed through a dual-language pipeline with language-specific routing: Fongbe videos transcribed with our fine-tuned MMS model, Hausa videos with NCAIR Whisper. Each segment was processed and returned with confidence scores. The output is a semi-supervised dataset of approximately 6,770 audio-transcription pairs in Parquet format with segment-level metadata (video ID, timestamp, confidence, domain label).

\subsection{Evaluation}

\paragraph{Curated data.} We report WER and Character Error Rate (CER) against reference transcripts using the \texttt{jiwer} library with standard tokenization. CER is critical for Fongbe given its 69-character diacritic-rich alphabet, as single diacritic errors can change word meaning.

\paragraph{YouTube data.} Without ground truth transcriptions, we employ three validation approaches: (1)~\textbf{Manual evaluation}---we randomly sampled 50 transcribed segments per language and had native speakers rate transcription quality on a 0--100 scale, where 0 indicates completely incorrect transcription and 100 indicates perfect transcription. Raters considered both semantic accuracy (correct meaning) and lexical accuracy (correct words and diacritics). (2)~\textbf{Confidence analysis}---we analyze the distribution of model confidence scores as a quality proxy. (3)~\textbf{Language identification}---we apply the \texttt{langid.py} toolkit to verify predicted language matches expected language, detecting potential code-switching or misclassification.

% ============================================================
\section{Results}
\label{sec:results}
% ============================================================

\subsection{Fongbe ASR Performance}

Table~\ref{tab:fon_results} presents results on the ALFFA test set (2,168 utterances, with diacritics preserved). Our fine-tuned MMS-300M achieves 9.48\% WER and 3.96\% CER, representing a 78\% relative WER reduction from the 2016 baseline. To our knowledge, this is the best reported result on this benchmark under diacritic-preserving evaluation.

\begin{table}[ht]
  \centering
  \caption{Fongbe ASR results on ALFFA test set (with diacritics preserved).}
  \label{tab:fon_results}
  \begin{tabular}{lcc}
    \toprule
    \textbf{Model} & \textbf{WER (\%)} & \textbf{CER (\%)} \\
    \midrule
    \citet{laleye2016fongbe} (2016, Kaldi GMM/HMM) & 44.04 & --- \\
    MMS-300M pretrained \citep{mms}                 & 23.7  & 9.2  \\
    MMS-300M-Fongbe (ours)                          & \textbf{9.48}  & \textbf{3.96} \\
    \bottomrule
  \end{tabular}
\end{table}

Training converged over 30 epochs: validation WER decreased from 60.2\% (epoch 3) to 10.9\% (epoch 28), with training loss dropping from 26.4 to 0.26. The low CER confirms effective preservation of Fongbe-specific characters and tonal diacritics. Qualitative inspection shows near-perfect transcription for standard utterances, with errors limited to minor function-word substitutions and occasional tone mark confusion on longer, complex sentences.

\subsection{YouTube Transcription Quality}

\begin{figure}[ht]
  \centering
  \includegraphics[width=\linewidth]{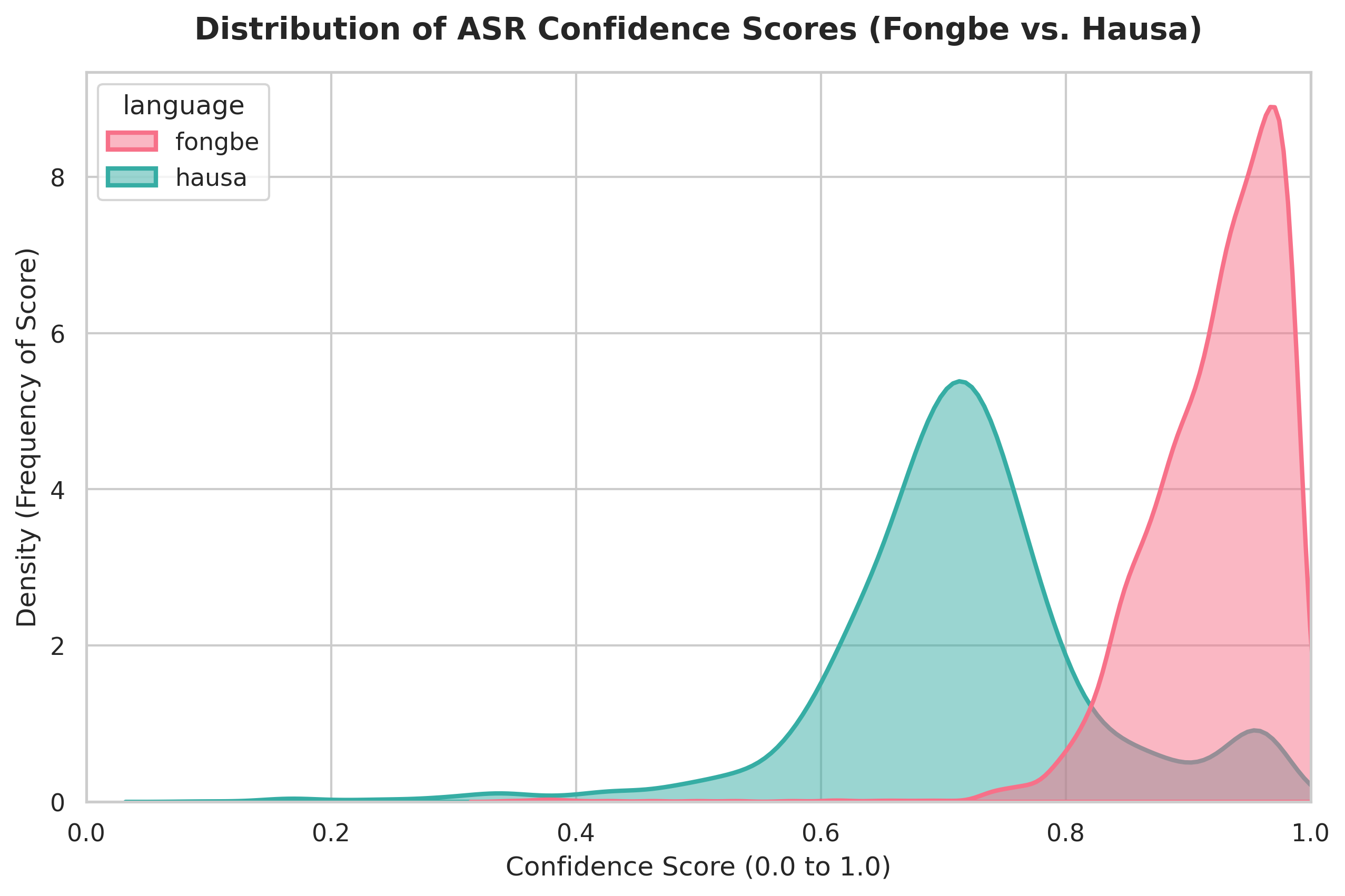}
  \caption{Distribution of ASR confidence scores for transcribed YouTube segments. The fine-tuned MMS-300M model for Fongbe shows high confidence (mean=0.84, $\sigma$=0.13) with 81\% of segments exceeding the 0.7 threshold, while the Hausa Whisper-Small model exhibits a broader distribution (mean=0.73, $\sigma$=0.19) with 65\% exceeding 0.7, likely reflecting higher noise levels and domain diversity in the Hausa video corpus.}
  \label{fig:confidence_plot}
\end{figure}

The pipeline produced 6,770 transcribed segments across both languages from the 424 selected videos (45.49 hours: 24.91h Fongbe, 20.57h Hausa). Figure~\ref{fig:confidence_plot} shows confidence score distributions, with Fongbe achieving higher mean confidence (0.84 vs.\ 0.73). The Fongbe subset consists primarily of educational content with controlled recording environments, while the Hausa corpus spans more diverse domains including outdoor news broadcasts and music performances, which may contribute to the difference in confidence distributions.

Table~\ref{tab:manual_eval} presents human evaluation results. Native speakers rated transcription quality on a 0--100 scale considering both semantic and lexical accuracy. Hausa achieves substantially higher quality (mean 57.4) with 60\% of segments rated as acceptable ($\geq$60), while Fongbe shows lower quality (mean 36.5) with only 20\% acceptable. Language identification accuracy remains high (94\% Fongbe, 89\% Hausa), indicating our language-specific models successfully avoid the French/Arabic misclassification issues observed in preliminary experiments with generic Whisper.

\begin{table}[ht]
  \centering
  \caption{Human evaluation on 50 random YouTube segments per language. ``$\geq$60'' = fraction rated acceptable. ``LID'' = language identification accuracy.}
  \label{tab:manual_eval}
  \begin{tabular}{lccccc}
    \toprule
    \textbf{Language} & \textbf{Mean} & \textbf{Median} & \textbf{Std} & \textbf{$\geq$60} & \textbf{LID} \\
    \midrule
    Fongbe ($n$=50) & 36.5 & 37.5 & 16.9 & 20\% & 94\% \\
    Hausa ($n$=50)  & 57.4 & 60.0 & 21.5 & 60\% & 89\% \\
    \bottomrule
  \end{tabular}
\end{table}

The performance gap between languages reflects the challenges of in-the-wild transcription for tonal languages: Fongbe's tone-marking requirements, combined with code-switching and acoustic variability in YouTube content, result in errors that affect both semantic and lexical accuracy. The Hausa model benefits from more robust pretraining data and simpler phonology, enabling better generalization to uncurated audio. While Hausa transcriptions approach usable quality for corpus construction (median 60/100), Fongbe transcriptions would benefit from post-processing or additional model improvements.

Domain inspection reveals quality variation by content type: educational Fongbe content (language lessons, tutorials) shows higher scores, while music videos with heavy instrumentation show lower quality. For Hausa, news content from BBC Hausa and VOA Hausa exhibits higher quality, while music and entertainment content shows more variable scores. These patterns align with confidence score distributions and acoustic expectations (clearer speech, less background noise in educational and news contexts).

Common error patterns include diacritic confusion for Fongbe (particularly tone marks), code-switching with French (Fongbe) and English (Hausa), and proper noun recognition challenges in both languages. For Fongbe, tone errors substantially affect semantic accuracy, explaining the lower human ratings despite high model confidence scores.

% ============================================================
\section{Discussion}
\label{sec:discussion}
% ============================================================

\paragraph{Viability of ASR-based corpus expansion.}
Our results reveal different viability outcomes for the two languages. For Hausa, ASR-based corpus expansion is viable: off-the-shelf fine-tuned models achieve 57.4/100 mean quality on in-the-wild audio, with 60\% of segments rated as acceptable ($\geq$60), indicating the transcriptions can support corpus construction with modest post-processing. For Fongbe, the picture is more complex: while language-specific fine-tuning achieved excellent performance on curated data (9.48\% WER, 78\% reduction from baseline), in-the-wild transcription quality is lower (36.5/100 mean, 20\% acceptable). The high model confidence (mean 0.84) masks semantic errors from incorrect tone marks, revealing a limitation of confidence-based quality estimation for tonal languages. Fongbe transcriptions in their current form would require substantial post-processing or additional model improvements before use in corpus construction.

\paragraph{The 12-hour sufficiency finding.}
A key result is that 12.3 hours of curated data was sufficient to fine-tune a 300M-parameter CTC model to competitive Fongbe performance, achieving a 78\% relative improvement over prior baselines. This finding is practically significant: for extremely low-resource African languages, collecting 10--15 hours of quality transcribed speech is feasible through small academic or community efforts, whereas accumulating 100+ hours requires institutional resources often unavailable. Our result suggests that modest data collection can bootstrap viable ASR pipelines, enabling the YouTube-to-text corpus expansion we demonstrate here.

\paragraph{Domain and content type matter.}
Educational and news content produces higher-quality transcriptions than music or cultural programming, as reflected in both confidence score distributions and human evaluation. This aligns with acoustic expectations (clearer speech, less background noise) but has important implications for corpus construction: prioritizing certain domains yields better text resources per compute hour. Future work expanding transcription coverage should consider domain quality when prioritizing additional videos from the catalog.

\paragraph{The confidence-quality gap for tonal languages.}
A critical finding is that model confidence scores are unreliable quality indicators for tonal languages in uncurated settings. Fongbe transcriptions achieved high model confidence (mean 0.84, with 81\% exceeding 0.7 threshold) but low human quality ratings (mean 36.5/100), revealing that the model confidently produces phonetically plausible but semantically incorrect outputs when tone marks are wrong. This confidence-quality gap does not appear for Hausa, where confidence and quality align more closely. For tonal low-resource languages, human evaluation or specialized tone-aware quality metrics are necessary to validate ASR output, as standard confidence scores may overestimate usability.

\paragraph{Limitations.}
Human evaluation was limited to 50 segments per language due to annotator availability; larger-scale evaluation would provide more robust quality estimates. We lack domain-stratified evaluation, preventing precise quality characterization by content type. We processed 424 videos from our catalog of 1,553 due to computational resource constraints; extending transcription to additional videos remains future work. The Fongbe training set (12.3h) is relatively small; incorporating additional data sources could improve in-the-wild performance, though our results suggest tone-aware training objectives or architectures may be necessary beyond simply adding more data. We did not fine-tune the Hausa model ourselves, limiting our ability to optimize for YouTube-specific acoustic conditions. The YouTube catalog is biased toward online content creators and may not represent offline speech patterns or regional dialects. The relatively low Fongbe quality scores (mean 36.5/100) indicate current transcriptions are not production-ready for corpus construction without post-processing.

% ============================================================
\section{Conclusion}
\label{sec:conclusion}
% ============================================================

We investigated ASR-based text acquisition for low-resource Fongbe and Hausa. Fine-tuning MMS-300M on 12.3 hours of curated Fongbe data achieved 9.48\% WER, the best reported result on the ALFFA benchmark under diacritic-preserving evaluation, demonstrating that modest data collection efforts can bootstrap competitive ASR for extremely low-resource tonal languages on curated benchmarks. However, applying these models to 424 videos from a catalog of 1,553 videos (236 hours total) revealed a critical gap between curated and in-the-wild performance: while Hausa transcriptions achieved moderate quality (mean 57.4/100, 60\% acceptable), Fongbe transcriptions scored substantially lower (mean 36.5/100, 20\% acceptable), indicating additional work is needed for production-ready tonal language ASR in uncurated settings.

Our work establishes four key findings: (1)~12 hours of curated data suffice to achieve competitive ASR performance on curated benchmarks for extremely low-resource tonal languages (9.48\% WER), but this does not guarantee in-the-wild quality; (2)~model confidence scores are unreliable quality indicators for tonal languages, with high confidence masking semantic errors from tone mark mistakes; (3)~ASR-based corpus expansion is viable for Hausa (non-tonal, better-resourced) but requires further development for Fongbe (tonal, extremely low-resource); (4)~domain-aware video selection matters, with educational and news content producing higher quality than music or entertainment.

We release four resources upon acceptance: a curated tone-preserved Fongbe speech dataset (13,581 utterances, 12.29h), a fine-tuned Fongbe ASR model, a YouTube catalog with 1,553 video IDs and metadata, and a transcribed corpus of 6,770 segments with human quality ratings.

\paragraph{Future work.} We will explore tone-aware ASR architectures or training objectives for improved in-the-wild Fongbe performance, post-ASR error correction for tone marks, confidence estimation methods sensitive to tone errors, and expansion to the full video catalog for Hausa (where quality is sufficient) while developing improved methods for Fongbe.

% ============================================================
\section*{Ethics Statement}
% ============================================================

Our YouTube data collection and release follow platform Terms of Service and ethical research guidelines. We catalog publicly available videos and provide video IDs, timestamps, and processing scripts rather than redistributing raw audio, consistent with fair use for academic research. The transcribed corpus includes only ASR-generated text and metadata, not speaker identities. An opt-out mechanism will be provided at release for content creators who wish to have their video IDs removed from the catalog.

% ============================================================
\bibliographystyle{plainnat}
\bibliography{asr_paper}
% ============================================================

\end{document}